**Type of manuscript:** Original research article

**Title:** A chart review process aided by natural language processing and multi-wave adaptive sampling to expedite validation of code-based algorithms for large database studies

**Contributors:** Shirley V Wang[1], Georg Hahn[1], Sushama Kattinakere Sreedhara[1], Mufaddal Mahesri[1], Haritha S. Pillai[1], Rajendra Aldis[1], Joyce Lii[1], Sarah K. Dutcher[2], Rhoda Eniafe[2], Jamal T. Jones[2], Keewan Kim[2], Jiwei He[2], Hana Lee[2], Sengwee Toh[3], Rishi J Desai[1*], Jie Yang[1*]

**Affiliations:**

[1] Division of Pharmacoepidemiology and Pharmacoeconomics, Department of Medicine, Brigham and Women's Hospital, Harvard Medical School, Boston, MA

[2] Center for Drug Evaluation and Research, Food and Drug Administration, Silver Springs, MD

[3] Department of Population Medicine, Harvard Medical School and Harvard Pilgrim Health Care Institute, Boston, MA

*Co-senior authors

**Corresponding author:**

Shirley V Wang

1620 Tremont St Suite 3030

Boston, MA 02120

Email: swang1@bwh.harvard.edu

Phone: 617-525-8376

**Conflict of Interest:** SVW has been an ad hoc consultant to Exponent Inc, Cytel Inc, and MITRE, a federally funded research and development center for the Centers for Medicare and Medicaid Services. ST is a methods consultant for Pfizer, Inc. Other authors do not have any conflicts of interest.

**Source of funding:** This project was supported by Task Order 75F0123F19010 under Master Agreement 75F40119D10037 from the US Food and Drug Administration (FDA). FDA coauthors reviewed the study protocol, statistical analysis plan, and the manuscript for scientific accuracy and clarity of presentation. Representatives of the FDA reviewed a draft of the manuscript for presence of confidential information and accuracy regarding statement of any FDA policy. The views expressed are those of the authors and not necessarily those of the US FDA.

**Data and computing code availability**: This methods study did not evaluate a hypothesis and therefore was not pre-registered. The study protocol with logged amendments, open-source SAS code in the CIDA query package used to extract the cohort of patients with obesity, and input data with accompanying R code used to calculate performance metrics and prepare figures are available in the project repository on the Open Science Framework: https://osf.io/3fe25/. The open-source CORA tool used for NLP aided chart review is available https://github.com/jiesutd/CORA. Our data use agreements prohibit sharing of source data and data derivatives with non-covered individuals and organizations. However, Medicare and Medicaid data may be requested through ResDAC: https://resdac.org/.

**Abstract:**

**Background:** One of the ways to enhance analyses conducted with large claims databases is by validating the measurement characteristics of code-based algorithms used to identify health outcomes or other key study parameters of interest. These metrics can be used in quantitative bias analyses to assess the robustness of results for an inferential study given potential bias from outcome misclassification. However, extensive time and resource allocation are typically required to create reference-standard labels through manual chart review of free-text notes from linked electronic health records.

**Methods**: We describe an expedited process that introduces efficiency in a validation study using two distinct mechanisms: 1) use of natural language processing (NLP) to reduce time spent by human reviewers to review each chart, and 2) a multi-wave adaptive sampling approach with pre-defined criteria to stop the validation study once performance characteristics are identified with sufficient precision. We illustrate this process in a case study that validates the performance of a claims-based outcome algorithm for intentional self-harm in patients with obesity.

**Results**: We empirically demonstrate that the NLP-assisted annotation process reduced the time spent on review per chart by 40% and use of the pre-defined stopping rule with multi-wave samples would have prevented review of 77% of patient charts with limited compromise to precision in derived measurement characteristics.

**Conclusion:** This approach could facilitate more routine validation of code-based algorithms used to define key study parameters, ultimately enhancing understanding of the reliability of findings derived from database studies.

**Background**

Studies that use large databases containing patient health information collected as part of routine care (e.g. claims, electronic health records [EHR]) are increasingly being used to generate real-world evidence to inform healthcare decision-making.[1-4] However, such data are collected for purposes other than research. Because of this, database studies often rely on clinical code-based algorithms to identify key study parameters, such as the health outcome of interest. The ability of such algorithms to accurately capture the clinical concept of interest can affect the validity of the research findings.[5] Therefore, it is important to conduct chart validation studies, first to understand an algorithm's measurement characteristics (e.g., positive predictive value [PPV]) by comparing against a reference standard, and second to use the measurement characteristics in quantitative bias analyses that evaluate the potential impact of bias due to misclassification.

Accurately identifying and characterizing patients who may or may not have a particular clinical condition (e.g., who experience a health outcome of interest) typically requires a chart validation study with detailed manual review of clinical notes to establish reference-standard labels in validation studies. The substantial human effort required for this step makes the process of conducting validation studies time-consuming and resource-intensive, ultimately hampering scalability. In certain applications, such as post-marketing safety surveillance of newly marketed medications, efficient completion of validation studies is a critical step to assess whether the data source being considered is fit-for-use.[6]

To that end, our objective was to propose a process that has the potential to substantially expedite the conduct of validation studies by leveraging Natural Language Processing (NLP)[7,8] and multi-wave adaptive sampling[9,10] methods. NLP techniques can expedite chart review by directing the attention of chart annotators to relevant clinical information highlighted in EHR free-text notes. Multi-wave adaptive sampling techniques that identify sequential samples of charts for manual review across strata of a population allow investigators to estimate performance characteristics other than PPV, while also setting predefined stopping boundaries for success and futility in terms of algorithm performance for a primary metric of interest, thus limiting the number of charts that need to be reviewed. By reducing the time and resources required for creating reference standard labels, incorporating these methods into an expedited chart review process could allow for more frequent and systematic validation of algorithms in large database studies, ultimately enhancing understanding of the reliability of findings derived from database studies.

**Methods**

There are seven iterative steps in the proposed chart review process flow chart (**Figure 1**). We describe the methodology for this process and illustrate the implementation via a case study focused on validation of a claims-based algorithm for intentional self-harm in adult patients with obesity. A study protocol with logged amendments is available in *e-supplemental materials 1*.

**Case study: Validation of claims-based algorithm for intentional self-harm in patients with obesity**

Suicidal outcomes are an important drug safety concern that clinical trials are generally not powered or designed to detect. Large healthcare database studies can generate complementary evidence on safety endpoints; however, accurate assessment of suicidal outcomes — including suicidality, suicidal ideation, attempted suicide, and intentional self-harm — are challenging to define and accurately capture in administrative claims data. A systematic review identified 34 validation studies for algorithms to capture suicidal outcomes and classified them into two types of algorithms. The first type included "screening" algorithms with higher sensitivity and lower specificity that were intended to be followed up with chart validation. The second type included "outcome definition" algorithms with more specific and restrictive criteria resulting in higher PPV that would be considered adequate to be used for a study endpoint.[11] Screening algorithms often included diagnosis codes for probable/possible cases such as poisoning or overdose without restrictions to specific healthcare settings. These screening algorithms generally had PPVs less than 60% whereas the outcome definition algorithms that had the highest PPVs (ranging from 74% to 92%) incorporated information on emergency department or hospital stays with specific combinations of diagnostic codes. Generally, the sensitivity, specificity, and negative predictive value (NPV) of the algorithms were infrequently reported.

For our case study, we validated a screening algorithm for intentional self-harm that included International Classification of Diseases, 10th Revision, Clinical Modification (ICD-10-CM) codes observed in any care setting and any diagnosis position (*e-supplemental materials 2*) in a study population of adult patients with obesity. This population is at higher risk for intentional self-harm than the general population. [12,13]

**Data Source**

Mass General Brigham (MGB) is one of six sites in the Sentinel Real World Evidence Data Enterprise (RWE-DE) that is frequently leveraged by the FDA for methodological and regulatory studies of interest.[14] MGB has linked Medicare claims data (2007-2020) and Medicaid claims (2000-2018) deterministically to EHRs for patients receiving care at the MGB healthcare system (linkage success rate 99.9% for Medicare and 99.3% for Medicaid). The claims data include demographics, enrollment start and end dates, outpatient dispensed medications, medical diagnoses and performed procedures across care settings including inpatient, outpatient, skilled nursing and rehabilitation claims. These are linked with clinical assessment files, including the minimum data set (MDS) of a comprehensive assessment of all residents in Medicare or Medicaid certified facilities, the Outcome and Assessment Information Set (OASIS), a patient assessment tool used in Medicare home health care to plan care, determine reimbursement, and measure quality, and the Inpatient Rehabilitation Facility Patient Assessment Instrument (IRF-PAI) which is used to collect patient assessment data for quality measure calculation. MGB is the largest healthcare provider system in Massachusetts, comprising >40 healthcare facilities across the full care continuum. The EHR databases contain information on patient demographics, medical diagnosis and procedures, medications, vital signs, smoking status, body mass index (BMI), immunizations, laboratory data, clinical notes, and reports. MGB investigators have direct access to full-text clinical notes for these patients, including ambulatory notes, discharge summaries, and specialty reports (e.g., psychiatry, pulmonary). The linked claims-EHR dataset includes a total of 1.2 million patients; of whom nearly 40% (~484,000) have at least one healthcare encounter in the most recent year. Finally, these patients have been deterministically linked to state vital statistic files from Massachusetts, Connecticut, and Vermont to track long term outcomes of all-cause and cause-specific mortality.

**Study Population**

The study population in which we validated the algorithm included patients over 18 years of age with obesity who were enrolled in Medicare or Medicaid, received care at MGB. The cohort entry date was the first date for which a patient had a relevant obesity diagnosis code and also met the eligibility criteria. Key eligibility criteria included a requirement for baseline observability within claims data (continuous enrollment for the 180 days prior to and including the cohort entry date) as well as EHR data (at least one visit within 180 days and at least one visit more than 180 days prior to the index date) and no evidence of self-harm diagnosis codes within the 180 days prior to and including the index date (additional eligibility criteria described in **Figure S1, Table S1**). Patients were allowed to enter the cohort after October 1st, 2016, in order to ensure that the baseline assessment windows were entirely within the era after the United States switched from ICD-9-CM to ICD-10-CM. Patients were followed until they had an intentional self-harm outcome, death, disenrollment from claims, or end of available data. The SAS based query package using the Sentinel Cohort Identification and Descriptive Analysis Tool (CIDA)[15] including code lists are available in *e-supplemental materials 3*.

**Chart Review Process**

1. *Create annotation guide with objective operational criteria*

   First, an annotation guide with objective criteria for defining the outcome of interest, for intentional self-harm, was created. The annotation guide is available in *e-supplemental material 4*. Annotators were instructed to look for positive assertions of suicide attempt or intentional self-harm in EHR clinical notes that referred to the time frame of 30 days prior to or on the date of relevant diagnosis. Notes after the date of relevant diagnosis could be used if the notes referred to an event that occurred within 30 days prior to or on the date of relevant diagnosis. Whenever available, the annotators relied on the psychological or psychiatric assessment of the clinical team treating the patient, for example, whether a 1:1 sit-in was ordered or there was a referral for psychiatric care.

2. *Classify endpoint as occurring (y = 1) or not (y = 0) during follow-up according to the claims algorithm*

   We used the CIDA query package and claims-based ICD-10-CM screening algorithm for intentional self-harm to identify outcomes for the cohort of patients with obesity.

3. *Create strata for sampling*

   Our sampling strata were defined by whether there is evidence to support a possible intentional self-harm event from claims data and/or EHR data as shown in **Figure 2**. Sampling within such strata and then applying inverse sampling weights (described in step 6) allows estimation of measures of performance other than PPV, such as NPV, sensitivity, and specificity.[9]

   To create these strata, we first used the claims-based ICD-10-CM screening algorithm to classify patients as claims positive for having an intentional self-harm event (claims+) or claims negative (claims-) during follow up. We then used MetaMap[16] to filter through EHR notes for all patients in the cohort. We identified a list of concept unique identifiers (CUI) related to self-harm by searching the UMLS meta-thesaurus (https://uts.nlm.nih.gov/uts/umls/home). Our search terms included: "self-harm", "suicide". The list of CUIs (1725 for self-harm and 967 for suicide, 2,681 unique CUIs after removing duplicates from the intersection of the two lists, *e-supplemental materials 5*). Using the identified CUIs, patients with no notes containing any such CUI within their follow up were considered EHR negative (EHR-), meaning that we considered these patients to have no evidence to support occurrence of intentional self-harm. Those with at least one mention of such CUI in their notes, suggesting a possible occurrence of intentional self-harm, were considered EHR positive (EHR+) for the purpose of creating strata.

In **Figure 2**, patients who were both claims- and EHR- (group 1) were not sampled for annotation. Instead, the reference standard label was assumed to be negative for intentional self-harm. Patients with cause of death in state death records coded as suicide (group 3) were assumed to have experienced intentional self-harm and were also not sampled for annotation; instead, the reference standard label was assumed to be positive for intentional self-harm.

For patients who were either claims+ or [claims- and EHR+] (group 2), equal numbers of claims+ and claims- patients were sampled in successive waves for annotation. For the claims+ patients, this group was subdivided based on whether the patients had healthcare contact at MGB within +/- 60 days of the claims-based algorithm outcome date (**Figure 2** boxes d, e, f, and g). This subdivision was made because the annotators were instructed to make their assessments based on the content of clinical notes. As annotators would not be able to make assessments for patients not seen within MGB because of the absence of clinical documentation, we could not conduct validation for this subsample (**Figure 2**, box d and e). We made the simplifying assumption that the proportion of true positives out of all claims+ was the same for patients seen at MGB as for patients seen outside the health system. For patients who were claims- but EHR+ (**Figure 2**, boxes b and c), notes within +/- 60 days of the first note with a relevant CUI related to self-harm occurring within patient follow up were reviewed for annotation.

4. *Select a random sample batch of size k*

    In our example, there were 265 patients for whom the claims-based screening algorithm for intentional self-harm was positive and had healthcare contact within the MGB healthcare system around the date of the relevant claim (**Figure 2** boxes f and g). We identified an equal number of randomly sampled patients who were claims negative but had CUIs related to self-harm or suicide (**Figure 2** boxes b and c).

    We took sequential waves of random samples from the full set of 530 patient charts (265 claims+ and 265 claims-). In each wave, we took a new batch of size $k = 10$ for chart review until the pool of samples was depleted. Each batch included five claims+ and five claims- patient charts (a disproportionate stratified sample). Inverse sampling weights were used to correct the sampling process. In parallel work, we have demonstrated that this approach balances simplicity and efficiency compared to more complex sampling strategies in a multi-wave setting, across a variety of scenarios.[17]

    In order to compare our performance characteristics at the point of meeting a stopping criterion to what the performance would have been had we continued reviewing, we completed chart review on the entire sample of 530 charts.

5. *Trained annotators review and label patients as endpoint occurring (Y = 1) or not (Y = 0)*

We used Clinical Optimized Record Annotation (CORA), a time-contextualized NLP-assisted EHR review tool developed in Python with PyQt5[18], which is compatible with Linux, Windows and Mac operating systems. CORA loads both structured data and unstructured patient notes, aggregates them, and displays them in chronological order within a graphical user interface (GUI) *(e-supplemental materials 6)*. It highlights terms in unstructured notes based on a user-customized keyword list to enhance manual review efficiency. To tailor CORA to our endpoint of interest, we uploaded the previously identified 2,681 unique CUIs related to self-harm. Doing so allowed CORA to highlight text that matched the CUIs and expedite the chart review process for the annotators.

The pair of annotators had both medical training and master's degrees in public health. They were instructed to assign the intentional self-harm label based on the content of EHR notes and in accordance with the criteria outlined in the annotation guide. There was a multi-step process for training annotators. First, 30 charts were sampled for double annotation. Next, inter-rater reliability was assessed using Cohen's Kappa.[19-21] As specified in our protocol, if kappa >0.8 then we would move onto independent annotation of subsequent waves of sampled batches. If kappa <0.8 then double annotation would continue, and the kappa was re-calculated in sequential waves until it exceeded 0.8. Charts where the annotator was unsure were discussed with a practicing board certified psychiatrist and resulting decisions were used to update the annotation guide.

The time required for annotation of each patient chart was automatically recorded by CORA. After training was completed with high kappa observed, the annotators were asked to independently review sequential samples of batch of k = 10 charts loaded into CORA. In order to compare annotation time with and without NLP assistance, each annotator additionally reviewed a sample of 20 charts *without* NLP highlighted terms that had previously been reviewed by the other annotator *with* NLP highlighted terms. This provided us with a sample of 40 charts with recorded timing of review both with and without NLP assistance.

6. *Compute the quantity of interest and confidence bands*

The primary measure of interest for the claims-based screening algorithm for intentional self-harm was the PPV. However, based on the sampling strategy, we were also able to estimate NPV, sensitivity, and specificity.

The PPV was estimated using all accumulated reference standard labels at each successive wave of sampling. Bayesian credible intervals were used to quantify uncertainty, denoted as p. Bayesian credible intervals were computed with a Beta-Binomial model, given by a Beta(1,1) prior on the unobserved parameter p which is then updated once Binomial samples are observed. The Beta(1,1) prior represents our non-informative belief that the PPV can range anywhere from 0 to 1. Upon observing more data, we update our model, leading to a refinement of the PPV estimate. To be precise, after having observed *s* successes (events) among *k* samples (reviewed charts, our knowledge of the PPV is described by a Beta(1+*s*,1+*k-s*) posterior). To arrive at a credible interval, we compute the *α*/2-quantile $q_{α/2}$ and the (1-*α*/2)-quantile $q_{1-α/2}$ of the Beta(1+*s*,1+*k-s*) posterior, which then form the credible interval [$q_{α/2}$,$q_{1-α/2}$]. Note that no alpha spending is required in the Bayesian case. The Bayesian credible intervals are always computed with *α*=0.05.

When estimating the other measurement characteristics, due to unbalanced sampling fractions from claims+ and claims- patients, we are not able to simply calculate the metrics based on annotations of the sampled charts. Instead, we estimate the metrics after weighting based on the inverse sampling fraction to reflect estimates of the metric as if the whole cohort had been annotated. Groups 1 and 3 were considered fully annotated as being truly negative or positive for intentional self-harm, respectively, therefore patients in these groups have a sampling weight of 1.0. The weights for group 2 were determined by the sampling fractions for claims+ versus claims- patients sampled for annotation at the time that the stopping criterion for PPV was met. This analysis was performed using the statistical language R (details on confidence interval calculation provided in *e-supplemental materials 7, analytic code and data in e-supplemental data 1*).

7. *Iterate steps 5-6 until stopping criterion is met*

    Because our primary metric of interest was the PPV, our stopping rules focused on sequential evaluation of the PPV. Our stopping rule was defined by the lower bound of the Bayesian credible interval for the PPV being above 0.75 (indicating high accuracy of the algorithm) or the upper bound of the Bayesian credible interval being below 0.75 (indicating futility in achieving an acceptably high performance).

**Ethics review**

This Sentinel project is a public health surveillance activity conducted under the authority of the FDA and, accordingly, is not subject to Institutional Review Board (IRB) oversight. It received exemption from Mass General Brigham IRB on their independent review.

**Transparency and data sharing statement**

This methods study did not evaluate a hypothesis and therefore was not pre-registered. The study protocol with logged amendments, open-source SAS code in the CIDA query package used to extract the cohort of patients with obesity, and input data with accompanying R code used to calculate performance metrics and prepare figures are available in the project repository on the Open Science Framework: https://osf.io/3fe25/.

The open-source CORA tool used for NLP aided chart review is available https://github.com/jiesutd/CORA.

Our data use agreements prohibit sharing of source data and data derivatives with non-covered individuals and organizations. However, Medicare and Medicaid data may be requested through ResDAC: https://resdac.org/.

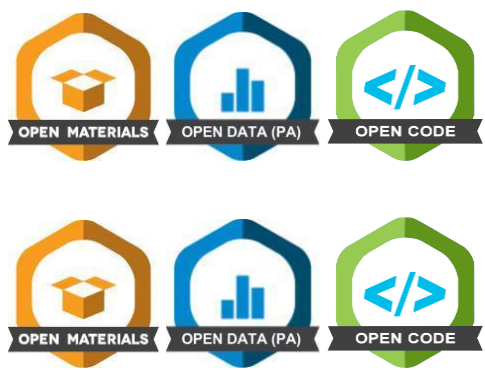

**Results**

There were 62,129 adult patients with obesity identified in the MGB EHR-claims database. The average age was 63.4 years, 62.1% of the patients were female, and 76.8% were white. The average BMI in this cohort was 36.3 kg/m$^2$. The prevalence of selected comorbidities, baseline medications, and healthcare utilization are shown in **Table 1**.

As the kappa between the two annotators was 100% after 30 charts, subsequent batches were independently annotated. The median time to review each patient chart using the NLP embedded in the CORA tool to highlight relevant text was 7.0 minutes. This ranged from less than 0.1 minute to 92.6 minutes. The average number of notes available to review per patient was 45.7 and ranged from 1 to 512 notes. During annotator training, which included dual review of the same 30 charts, the two annotators had similar timing in terms of review. When the trained annotators were asked to independently review the same charts, half *without* the use of NLP highlighted terms that the other annotator reviewed *with* NLP assistance, we observed that the median time to review for this subset of charts was 6.0 minutes for NLP aided annotation compared to 11.4 minutes for annotation without NLP assistance.

In randomly sampled sequential batches of k = 10 charts, the cumulative estimated PPV met the stopping criterion for futility after 12 batches (23% of the full sample). At this point, the estimated PPV was 0.60 with 95% credible interval (0.47, 0.72). As shown in **Figure 3**, the estimated cumulative PPV stabilized around the time of the stopping criterion being met. Annotating additional charts (total = 530) marginally tightened the credible intervals without substantive changes to the point estimate. The inverse sampling weighted sensitivity, specificity, and NPV are shown in **Table 2** and indicated generally high performance that were consistent at the time of meeting the stopping criterion and after review of the full sample, although confidence intervals were wide for sensitivity due to the large weights within the small sample.

We characterized the reasons why the annotators classified cases as false positives in **Table 3**. The most common reason was because the only text related to self-harm in the EHR notes were mental status assessments which indicated no suicidal or homicidal thoughts or behavior (28.6%). The next most common reason was that all of the text in the notes related to past history of suicide attempts or intentional self-harm, occurring more than 90 days prior to the relevant claims code (18.4%). Many of these events were documented as occurring years prior to the claims code. The third most common reason was due to patient drug overdose, where the patient denied that the overdose was a suicide attempt, and the psychological evaluation and subsequent treatment plans were aligned with this assessment of lack of intent to cause self-harm (16.3%).

Twenty-three patients were classified as false negatives because they were completed suicides identified by cause of death in linked state death records without recent healthcare contact that resulted in a relevant claims diagnosis code or CUI in an EHR note (**Figure 2**). There was also a false negative identified in the chart reviewed sample where there was evidence of a suicide attempt in the EHR notes, resulting in psychiatric hospitalization; however, upon review of claims diagnoses, only diagnosis codes related to suicidal ideation were found. Codes for suicidal ideation were not included in the intentional self-harm claims code algorithm.

**Discussion**

In this paper, we introduce a chart review process that leverages two distinct enhancements 1) NLP assisted review and 2) multi-wave adaptive sampling. The process is designed to expedite validation of code-based algorithms for large database studies and support sensitivity analyses that evaluate the potential impact of bias due to misclassification. In our illustrative case study that validates a screening algorithm for intentional self-harm in patients with obesity, we observed that using a stopping criterion for futility in the adaptive multi-wave samples could have saved substantial time and resources by preventing unnecessary review of 77% of charts. In addition, NLP aided review with the open-source CORA tool reduced the amount of time to annotate charts by 48% without compromising inter-annotator reliability.

The proposed process has many strengths. It can be flexibly applied across study populations and algorithms. The sampling strategy based on strata defined by sources of evidence makes it possible to estimate performance metrics other than PPV, such as sensitivity, specificity, and NPV. Alternative stopping rules can be applied, depending on the needs and priorities of the investigators (e.g., width of credible interval, success and futility set at different levels). CORA is open-source, lightweight, has an easy-to-use GUI, and is compatible with clinical notes extracted from different healthcare systems.

There are also limitations to the proposed process that merit discussion. First, we used a relatively simple NLP approach based on MetaMap to both identify "true negatives" and identify concepts to highlight in clinical notes. Although we cast a wide net to be sensitive in terms of relevant CUIs when filtering through notes, it is possible that patients whose notes do not include language that match the CUIs will be incorrectly classified as "true negatives". Additionally, while negation detection is conducted within MetaMap when searching for concepts, we should acknowledge that it may not be fully accurate. Some negated concepts may be incorrectly recognized as present, some affirmatively-mentioned concepts may be missed, and that certain words, phrasing, and clinical concepts may be missed or misclassified due to synonymy, the complexity of human language, and/or the unique semantic and structural challenges posed by clinical notes. As newer NLP approaches continue to develop, including large language models (LLM), they can be plugged into the framework of the proposed process for more sophisticated NLP-aided human review. However, in the context of chart review, the NLP is used only as an aid to direct the attention of the human annotator to areas where there is likely to be relevant text to review. That said, any NLP-assisted tool could have the effect of reducing annotator attention to areas that are not highlighted. A limitation of the stratified sampling approach to facilitate estimation of performance characteristics is the necessary assumption that the proportion of true positives is the same for the subset of patients who are seen within the healthcare system to which claims data are linked as it is for patients who are seen in other healthcare systems. In general, when using success/futility stopping criteria with NLP-aided annotation and multi-wave adaptive sampling, the greatest efficiency gains are anticipated for algorithms that are clearly in the high validity range or in the futility range. For algorithms that are near the border, gains may be more limited.

There are also limitations to chart review that are not resolved by this process. The assumptions that are used for quantitative bias sensitivity analyses are informed by the measurement characteristics reported from chart review studies. These assumptions are more realistic and useful for bias adjustment when measurement characteristics are available separately for the exposure and comparator group of interest.[23] In our illustrative example, we focused on a defined study population, agnostic to exposures. This is in line with practical application of an expedited chart review where the population and endpoint of interest are known, but either the exposure/comparator is not selected, the exposure/outcome combination is too rare, or the goal is to obtain a more generic evaluation of performance that could be flexibly applied in quantitative bias analyses across many potential exposures of interest.

Finally, self-harm and suicidal outcomes are challenging to define and different case definitions may be appropriate for different goals. There is not a uniform set of terms or taxonomy for identifying this spectrum of ideation and/or behaviors.[24] Our case definition was based on identifying *intentional* self-harm and inferring this intent based on the reactions of the treating clinical team as documented in the EHR notes. The estimated PPV was low but in the same range or higher than many other screening algorithms for intentional self-harm.[11] Alternative case definitions would result in different performance metrics (example in *e-supplemental appendix 8*).

**Conclusion**

We have demonstrated a proposed process to expedite conduct of validation studies through more efficient chart review. The time and resource savings from implementing this process could facilitate more routine validation of outcome and other algorithms that are used to define key study parameters and enhance understanding of the potential impact of bias due to misclassification in database studies. Future work could add to the capabilities of this process by building in iterative tuning of claims-based algorithms to optimize performance for a given case definition/phenotype over multi-wave adaptive samples, rapidly testing alternative case definitions, or incorporating LLMs that iteratively learn and enhance the usefulness of NLP aided highlights in successive waves.

**Tables and figures**

**Table 1.** Characteristics of cohort of patients with obesity**Table 2.** Performance characteristics at the time a stopping criterion was met versus after review of full sample

**Table 3.** Reasons for false positive and false negative annotation

**Figure 1.** Process diagram for natural language processing assisted chart review with adaptive, multi-wave sampling

**Figure 2.** Counts in stratified sampling for estimation of performance metrics

**Figure 3.** Positive predictive value in cumulative multi-wave samples

*Supplement*

**Figure S1.** Design diagram for obesity cohort

**Table S1.** Eligibility flow for obesity cohort

*Stored on OSF:*

**eSupplemental Materials 1.** Study protocol with logged amendments

**eSupplemental Materials 2**. International Classification of Diseases 10[th] Revision code list for intentional self-harm

**eSupplemental Materials 3.** SAS based query package using the Sentinel Cohort Identification and Descriptive Analysis Tool (CIDA)

**eSupplemental Materials 4.** Annotation guide

**eSupplemental Materials 5**. Concept Unique Identifiers related to intentional self-harm or suicide

**eSupplemental Materials 6**. Screenshot of graphical user interface (GUI) for Clinical Optimized Record Annotation (CORA).

**eSupplemental Materials 7.** Calculation of weighted confidence intervals

**eSupplemental Materials 8.** Positive predictive value with alternative case definition

**eSupplemental Data 1**. De-identified annotation data and analytic code

**Table 1. Characteristics of cohort of patients with obesity**

|  | MGB EHR-claims linked data (2016-2020) | |
|---|---|---|
| **Patient characteristics** | N/mean | %/Std deviation |
| Unique Patients | 62,129 | N/A |
| Age (Years) | 63.4 | 16.2 |
| Age category | | |
| 18-39 | 7519 | 12.1 |
| 40-64 | 16,009 | 25.8 |
| ≥65 | 38,601 | 62.1 |
| Sex | | |
| Female | 38,556 | 62.1 |
| Male | 23,573 | 37.9 |
| Race | | |
| American Indian/Alaskan Native | 64 | 0.1 |
| Asian | 303 | 0.5 |
| Black or African American | 4,587 | 7.4 |
| Multiracial | <11* | 0 |
| Native Hawaiian or Other Pacific Islander | <11* | 0 |
| White | 47,698 | 76.8 |
| Unknown | <11* | 15.2 |
| Hispanic | | |
| Yes | 2,359 | 3.8 |
| No | 52,654 | 84.7 |
| Unknown | 7,116 | 11.5 |
| Year of cohort entry | | |
| 2016 | 20,075 | 32.3 |
| 2017 | 13,566 | 21.8 |
| 2018 | 13,383 | 21.5 |
| 2019 | 9,101 | 14.6 |
| 2020 | 6,004 | 9.7 |
| Region | | |
| Midwest | 123 | 0.2 |
| Northeast | 59,824 | 96.3 |
| South | 1,900 | 3.1 |
| West | 241 | 0.4 |
| Other | 23 | 0 |
| Missing | 18 | 0 |

| Insurance | | |
|---|---|---|
| Medicaid | 10,280 | 16.5 |
| Medicare | 41,800 | 67.3 |
| Dual | 10,049 | 16.2 |
| Combined Comorbidity Score | 2.5 | 3.1 |
| Claims-based frailty index | 0.2 | 0.1 |
| Type 2 Diabetes Mellitus | 22,273 | 35.8 |
| Body Mass Index (BMI) in kg/m$^2$, mean (SD) | 36.3 | 25.8 |
| Dementia | 2,536 | 4.1 |
| Psychosis | 3,554 | 5.7 |
| Anxiety | 16,981 | 27.3 |
| Depression | 18,614 | 30 |
| Bipolar or schizophrenia | 5,051 | 8.1 |
| Alcohol abuse or dependence | 2,696 | 4.3 |
| Antidepressants | 24,227 | 39 |
| Anxiolytics or Hypnotics | 5,129 | 8.3 |
| Antipsychotics | 6,256 | 10.1 |
| Benzodiazepines | 13,941 | 22.4 |
| GLP-1 agonists prior use | 1,663 | 2.7 |
| GLP-1 agonists concurrent use | 1,300 | 2.1 |
| SGLT2 inhibitors prior use | 632 | 1 |
| SGLT2 inhibitors concurrent use | 507 | 0.8 |
| DPP4 inhibitors prior use | 1,383 | 2.2 |
| DPP4 inhibitors concurrent use | 1,124 | 1.8 |
| EHR continuity measure | 0.5 | 0.4 |
| Low Income Subsidy | 29,436 | 47.4 |
| Mean number of ambulatory encounters | 16.7 | 16.7 |
| Mean number of emergency room encounters | 1 | 2.2 |
| Mean number of inpatient hospital encounters | 0.5 | 1 |
| Mean number of non-acute institutional encounters | 0.6 | 3.7 |
| Mean number of other ambulatory encounters | 2.5 | 5.6 |
| Mean number of filled prescriptions | 22.5 | 19.9 |
| Mean number of generics dispensed | 8.7 | 5.4 |

MGB: Mass General Brigham
EHR: electronic health record
GLP-1 agonist: Glucagon-like peptide-1 agonist
SGLT2 inhibitor: Sodium-glucose cotransporter 2 inhibitor
DPP4 inhibitors: Dipeptidyl peptidase-4 inhibitor
*Not shown per CMS small cell suppression policy

**Table 2. Performance characteristics at the time a stopping criterion was met versus after review of the full sample**

|  | When stopping criterion met (N = 120) | After review of all charts (N = 530) |
|---|---|---|
| PPV[1] | 0.6034 (0.4744, 0.7193) | 0.6343 (0.5751, 0.6897) |
| NPV[2] | 0.9996 (0.9849, 0.9996) | 0.9988 (0.9948, 0.9995) |
| Sensitivity[2] | 0.9318 (0.2060, 0.9354) | 0.8158 (0.4899, 0.9187) |
| Specificity[2] | 0.9964 (0.9952, 0.9974) | 0.9968 (0.9963, 0.9973) |

PPV = positive predictive value
NPV = negative predictive value
[1] PPV reflects cumulative estimates over multiple waves of sampling; Values in parentheses represent 95% credible intervals
[2] NPV, Sens, Spec are inverse sample weighted to the cohort size/distribution (sensitive to weights in small samples). Values in parentheses represent 95% confidence intervals

**Table 3 Reasons for false positive annotation**

| False Positives | Total (N = 98) | |
|---|---|---|
| **No evidence of intentional self-harm or related concepts** | N | % |
|     The only text related to suicidal outcomes comes from mental status assessment (e.g. "no suicidal or homicidal behavior or thoughts") | 28 | 28.6% |
|     No EHR notes referencing events within the 30 days prior to the date of relevant claims diagnosis | 3 | 3.1% |
| **History of concepts related to intentional self-harm: ideation, family history, and history of drug overdose or alcohol withdrawal** | | |
|     Family history | 2 | 2.0% |
|     Past history of suicidal ideation | 1 | 1.0% |
|     History of drug overdose or alcohol withdrawal | 5 | 5.1% |
| **Timing of intentional self-harm outside of case definition** | | 0.0% |
|     Past history of suicide attempt/intentional self-harm >90 days prior to relevant claims diagnosis | 18 | 18.4% |
|     Recent history of suicide attempt/intentional self-harm within 31-90 days prior to relevant claims diagnosis | 7 | 7.1% |
|     Suicide attempt/intentional self-harm AFTER relevant claims diagnosis | 2 | 2.0% |
| **Evidence of suicidal ideation without evidence of self-harm** | | |
|     Active suicidal ideation | 4 | 4.1% |
|     Passive suicidal ideation | 12 | 12.2% |
| **Evidence of current overdose without intent to self-harm** | | 0.0% |
|     Current overdose but patient denies suicide attempt/intentional self-harm, psychological/psychiatric evaluation and clinical follow up consistent with this assessment | 16 | 16.3% |

**Figure 1. Process diagram for natural language processing assisted chart review with adaptive, multi-wave sampling**

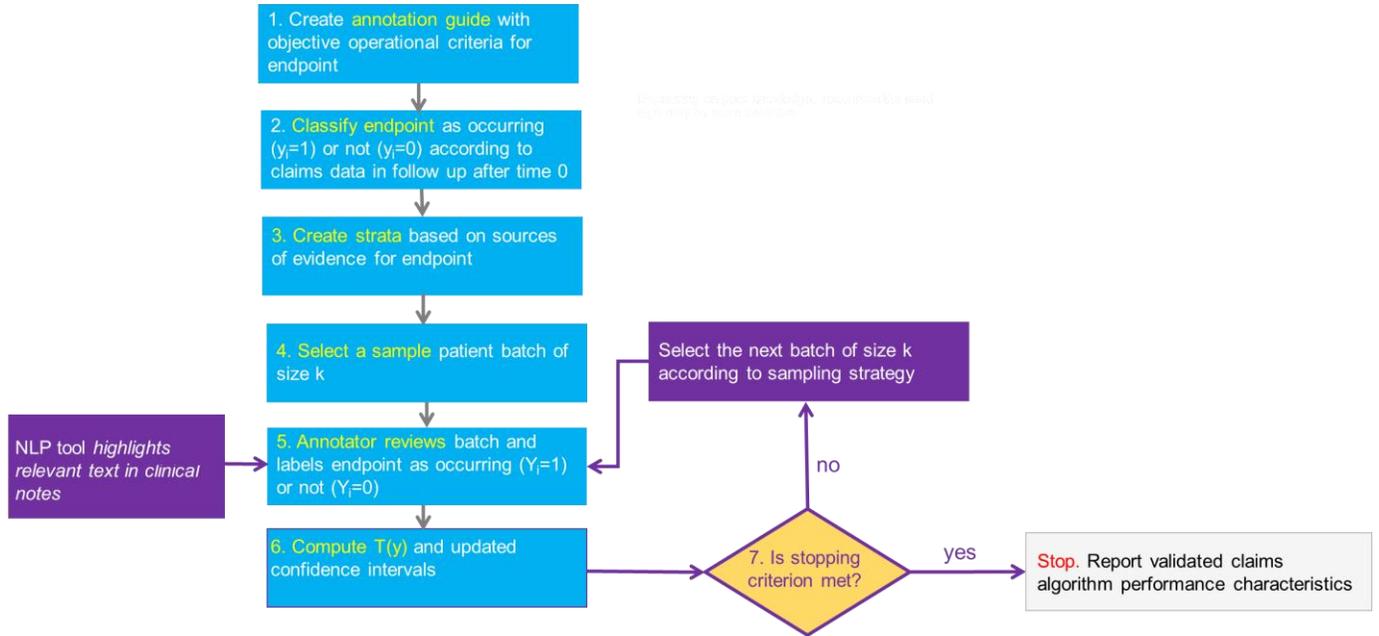

T(y) = measurement characteristic of interest

y = claims algorithm classification of the outcome (1 = outcome occurred, 0 = outcome did not occur)

Y = reference standard label for the outcome (1 = outcome occurred, 0 = outcome did not occur)

The NLP tool used in our example was Clinical Optimized Record Annotation (CORA): https://github.com/jiesutd/CORA

**Figure 2.** Counts in stratified sampling for estimation of performance metrics

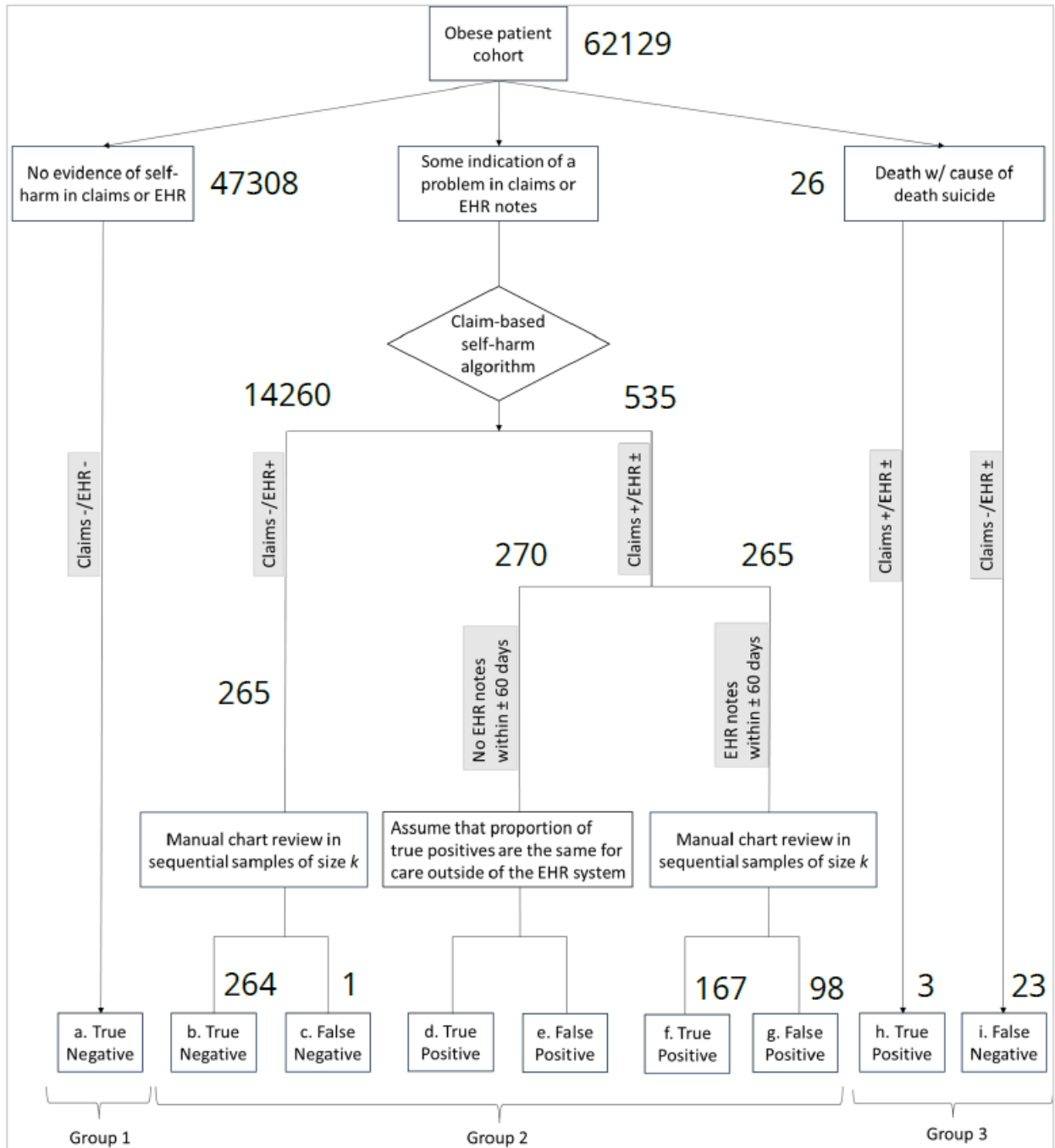

Boxes d and e are not reviewable due to lack of patient contact with the healthcare system where there is linked electronic health records to claims data.

**Figure 3. Positive predictive value in cumulative multi-wave samples**

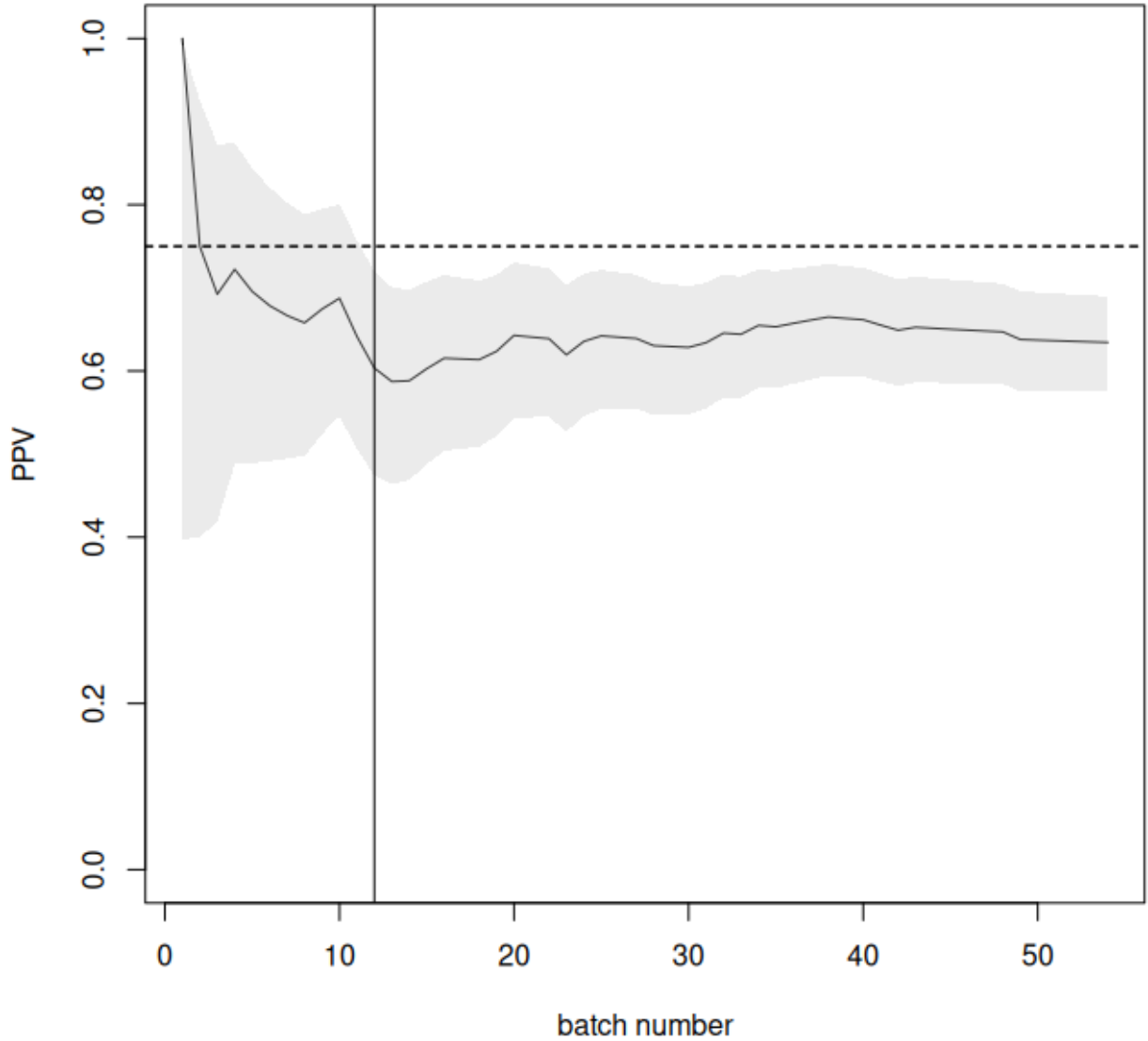

Dashed horizontal line is at the stopping criteria. "Success" stopping criterion = lower bound of credible interval exceeds 0.75. "Futility" stopping criterion = upper bound of credible interval below 0.75. Vertical solid line represents the point at which the stopping criterion was met. Each batch included 10 patient charts with at least 1 relevant claims diagnosis or CUI in the EHR notes.

**Figure S1. Design diagram for obesity cohort**

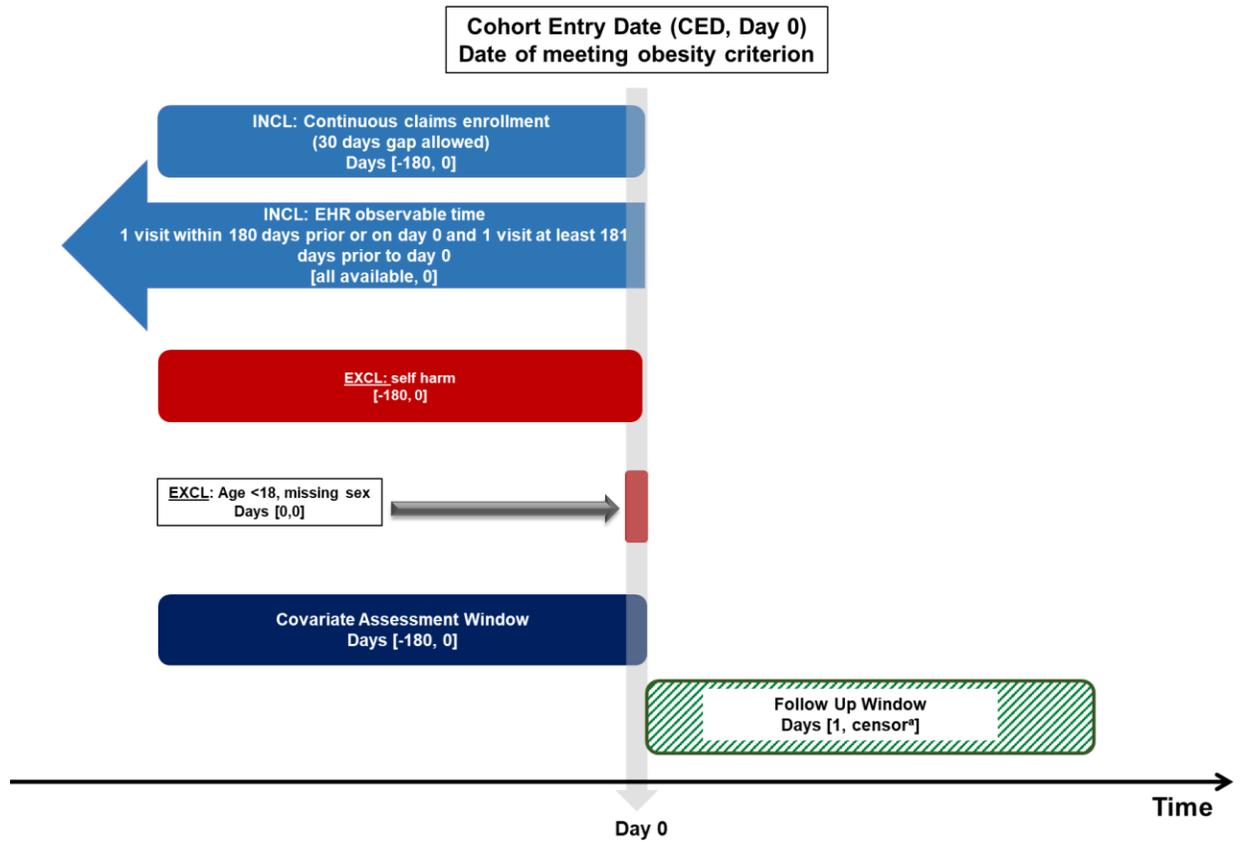

**Table S1. Eligibility flow for obesity cohort**

| Eligibility criteria | Obesity cohort |
|---|---|
| Output cohort from Query Request Package (QRP) | 134,019 |
| Exclude if patients do not have at least 1 EHR visit in baseline (180 days) | 66,838 |
| Exclude if patients do not have at least 1 EHR visit prior to baseline | 62,129 |
| Final obesity cohort | 62,129 |